# Loading Ceramics: Visualising Possibilities of Robotics in Ceramics

**Varvara Guljajeva**
Academy of Media Arts Cologne, Germany
varvarag@gmail.com

**Mar Canet Sola**
Baltic Film, Media and Arts School,
Tallinn University, Estonia
Academy of Media Arts Cologne, Germany
mar.canet@gmail.com

**Martin Melioranski**
Faculty of Architecture, Estonian Academy of Arts, Estonia
martin.melioranski@artun.ee

**Lauri Kilusk**
Faculty of Design, Estonian Academy of Arts, Estonia
lauri.kilusk@artun.ee

**Kaiko Kivi**
Faculty of Architecture, Estonian Academy of Arts, Estonia
kaiko.kivi@artun.ee

**Abstract**
This article introduces an artistic research project that utilises artist-in-residency and exhibition as methods for exploring the possibilities of robotic 3D printing and ceramics. The interdisciplinary project unites artists and architects to collaborate on a proposed curatorial concept and Do-It-With-Others (DIWO) technological development. Constraints include material, specifically local clay, production technique, namely 3D printing with a robotic arm, and kiln size, as well as an exhibition concept that is further elaborated in the next chapter. The pictorial presents four projects as case studies demonstrating how the creatives integrate these constraints into their processes. This integration leads to the subsequent refinement and customization of the robotic-ceramics interface, aligning with the practitioners' requirements through software development. The project's focus extends beyond artistic outcomes, aiming also to advance the pipeline of 3D robotic printing in clay, employing a digitally controlled material press that has been developed in-house, with its functionality refined through practice.

**Background: Estonian Academy of Arts clay printing facilities and their uniqueness**

3D printing with extrusion-based plastic materials commenced in 2006 at Estonian Academy of Arts (EKA) - Faculty of Architecture, 3D Lab (3DL), featuring the Stratasys Dimension SST 768 system for ABS plastic and proprietary soluble support material. In 2015, Faculty of Design and Department of Ceramics initiated the additive fabrication of clay and related material studies.

In 2017, 3DL received a UR10 collaborative robot, which further expanded the collaborative investigations between the two faculties. The establishment of the cross-faculty Protolab in 2018 was crucial for these developments. This lab has since evolved into a prominent centre specialising in the design and construction of custom CAD/CAM equipment. The collaboration among Protolab, 3DL, and the Ceramics Workshop enables concurrent develop-ment of goals, addressing distinct inquiries of the involved creative disciplines, and the requisite equipment, software, and materials, facilitating the realisation of intermediate stages and outcomes in the physical realm.

In order to overcome a limitation of maximum clay capacity that all available 3D clay printers have [1], an in-house Smart Press was developed. This special press is for depositing extrudable materials for the collaboration robot. The device has a unique design that allows continuous feeding for large-scale prints.

Regarding custom-made software to control clay printing machinery, a special middleware for operating UR10 as a 3D-printer in unison with both the Smart Press and Extruder was developed. The basis for software has been created primarily on McNeel Rhinoceros 3D and Grasshopper. Operation of the robot is assisted by a desktop server application built with NodeJS and help with ElectronJS platform providing the continuous movement code to the robot. Driver software for the Arduino platform running pump and extruder is developed in-house, too. This development has generated many opportunities to study the potential of the material and technology further with the development of necessary software.

### Loading Ceramics

"Keraamika kannab | Loading Ceramics" was a multifaceted initiative, encapsulating four distinct yet interconnected components: a short-term artist-in-residency program, hackathon, symposium, and exhibition. The selection of participants was based on their skills and motivation to explore and contribute towards advancing the manufacturing process with ceramic 3D printing. The initiative brought together a diverse group of nine talented individuals: five architects and four artists. Each participant was given the opportunity to bring their ideas to life, with the printing duration for each project ranging from two to five days, depending on the complexity and requirements of the project. Eight artist-in-residence shifts resulted in ceramic artefacts that were displayed in a group

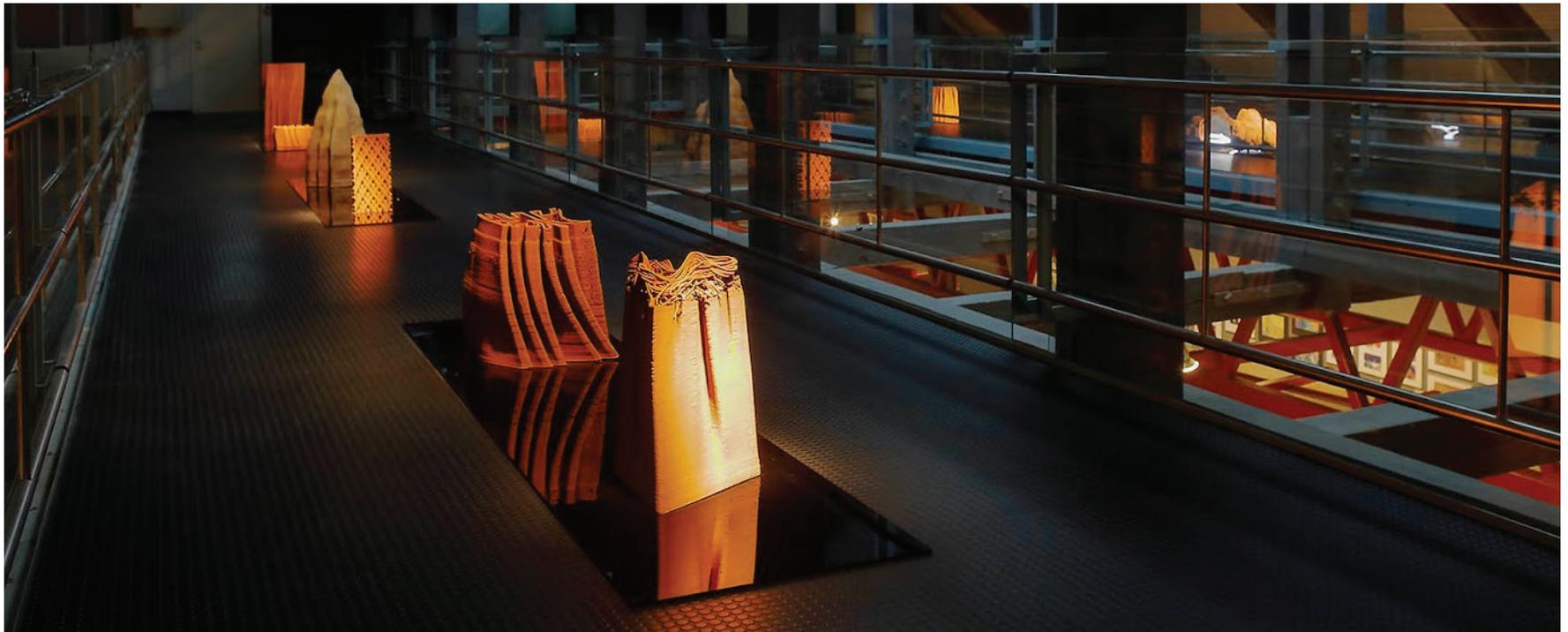

Exhibition "Keraamika kannab | Loading Ceramics" at athe Estonian Architecture Museum from 12 Sep - 12 Nov 2023. Photos by Juta Kübarsepp

exhibition "Keraamika kannab | Loading Ceramics" at the Estonian Architecture Museum from September 12th to November 12th, 2023.

The curatorial concept centred on re-envisioning the 'pillar' in architecture through collaborative robotic 3D printing of clay, examining its historical evolution as both a structural and cultural element. This exploration addressed the challenge of balancing minimization and maximisation: creating lightweight, easy-to-fabricate support structures while promoting a multidisciplinary synergy. The objective was to uncover and reinterpret the latent possibilities of support systems, often overlooked, by crafting experimental ceramic columns and pillars that resonate with current contexts.

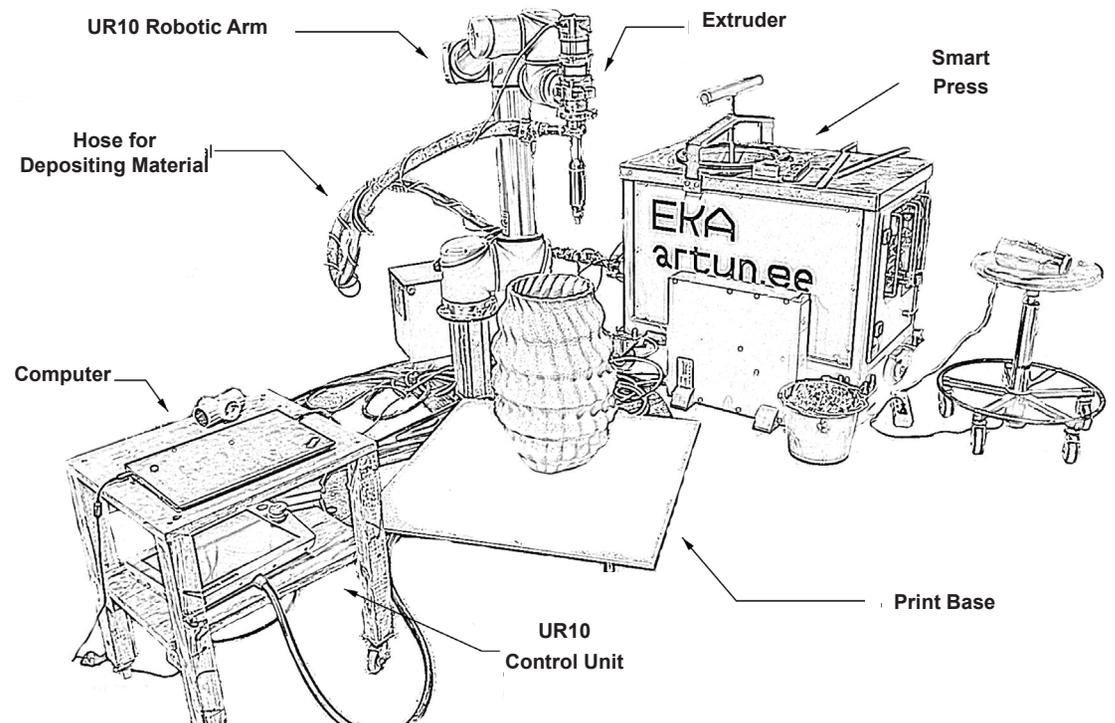

Skematic overwiew of the clay 3D-printing system components in EKA.

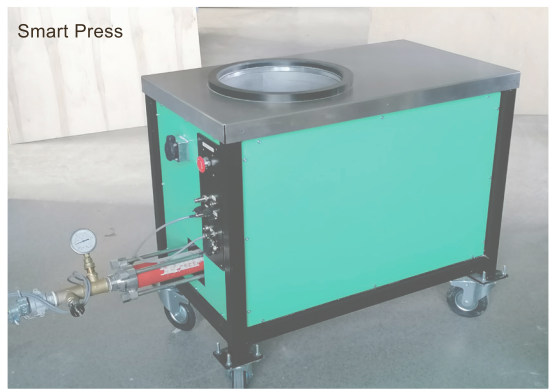
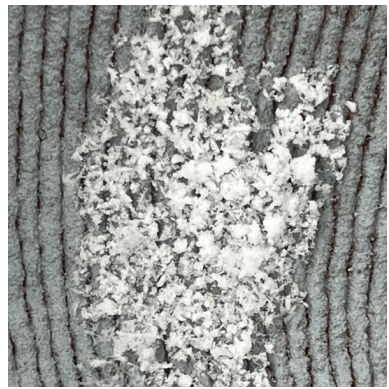
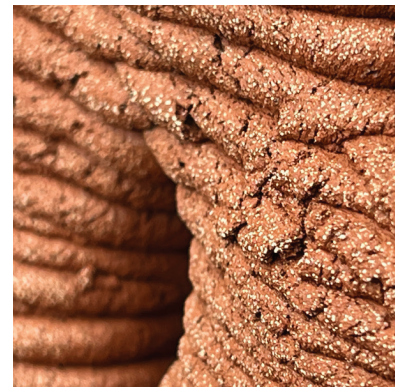
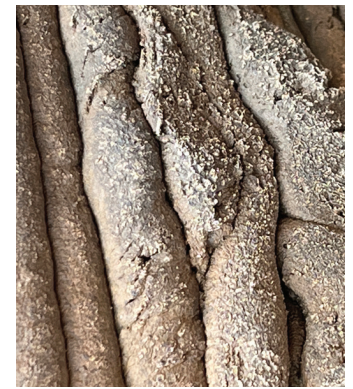

**Motivation and Research Methods**

The purpose of this research was to improve the workflow and machinery after initial trials, evaluating the viability of the innovative techniques for future creative developments. Focusing on the crucial physical properties of materials with plastic characteristics, the project examined their impact on the final print's structural and aesthetic attributes. Rooted in the Do It With Others (DIWO) approach, the endeavour promoted collaboration, collective knowledge exchange, artistic innovation, and technological progress, stemming from the artist-in-residence program. The term DIWO, first mentioned by Furtherfield in 2006, highlights the social aspect in comparison to the DIY movement, promoting radical openness, artistic freedom, and co-creation [2]. Hence, central to the research was technology decolonization and the synergy of creative minds, involving the development of synchronous pumps and control software by explorative professionals to meet their specific design goals.

In this research, we combined Research through Design (RtD) and artistic research methodologies. RtD focuses on the process and advancing design thinking, particularly in the context of 3D printing with clay. It is a method of inquiry in design practice that involves multiple iterations to arrive at an optimal solution [3]. Our research question centred on improving the current 3D printing setup for ceramics and exploring the new structures that could be invented by applying the DIWO method.

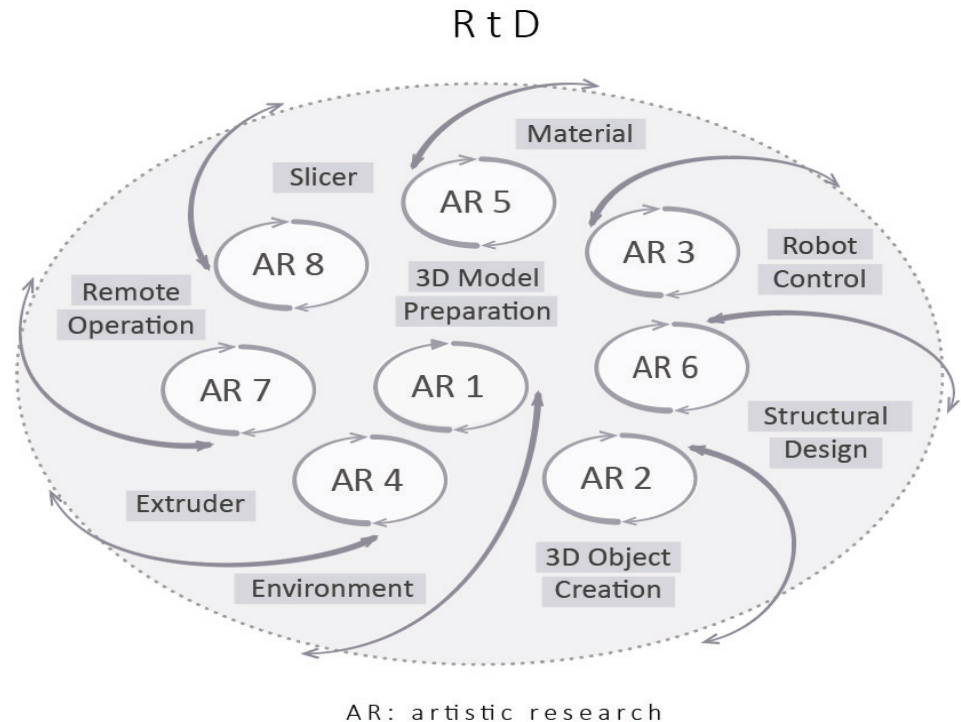

AR: artistic research

Through this combined approach, we aimed to push the boundaries of both the technological and artistic aspects of ceramic 3D printing.

The artistic research methodology, on the other hand, is concerned with the artistic process and final artefacts [4,5]. Each invited artist or architect engaged in practice-based research tailored to their individual projects, informed by the curatorial statement and constrained by material and fabrication technology. Simultaneously, they contributed to the RtD approach, which was a central objective of the DIWO segment of the research project.

This dual approach fostered an environment where participants not only developed their personal projects but also collaboratively enhanced the printing system and identified which structures proved effective and which did not. The insights and outcomes of this innovative exploration were visually documented and analysed in the subsequent four case studies.

Out of eight projects, we selected four to describe as case studies, detailing the process and methodology using a visual language of pictorial format. The authors explain their artistic research and elaborate on their contributions to the RtD method.

# Material Research in the Context of clay 3D-printing

During the "Loading Ceramics" residency program, all works were printed using locally sourced traditional brick-making material, which is naturally abundant throughout North Estonia. This earthenware clay dates back roughly 500 million years to the Cambrian era [6]. It is found exposed in the Baltic Glint along the northern coast of Estonia, situated between limestone layers. The main excavation site is located in the Kunda administrative unit at Aseri village, near a historical brick factory. The purified, dried, and sieved raw material is available as powder in local hardware stores.

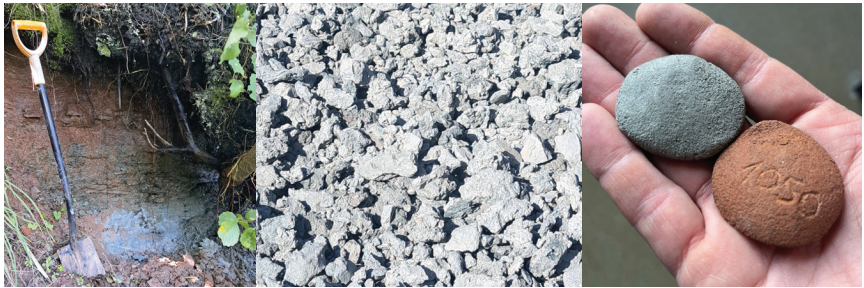

Natural location of the mineral deposit, sun-dried raw material, and material samples before and after firing.

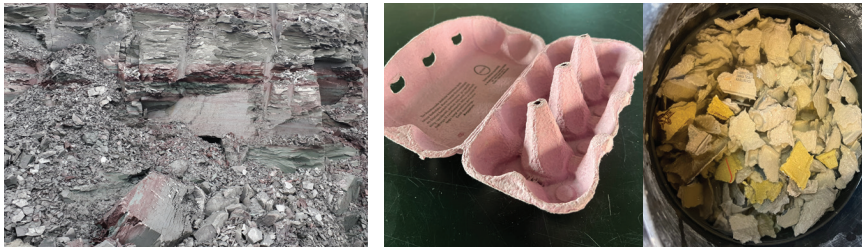

Clay settlements in Kunda quarry. Photo by Laura Põld.

Egg-cart before and after shredding and soaked in water for mixing.

For large-scale printing during this project, the clay powder was mixed with sand, recycled paper pulp, and water. Used egg cartons were first torn into pieces, soaked in water, and then blended into smaller particles. The paper pulp adds fiber to the filament, while the water allows for the regulation of the overall hardness and softness of the mix.

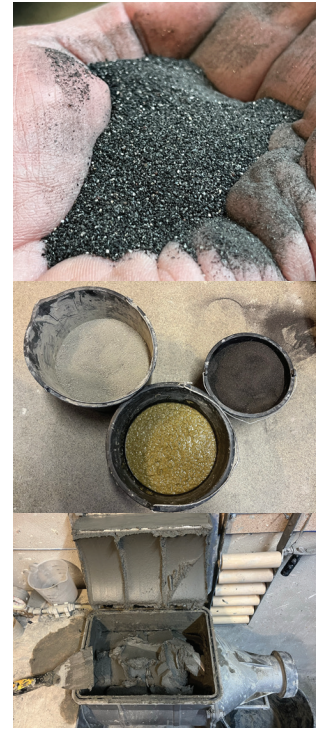

Sand acts as a stabiliser in the mix, reducing shrinkage and deformations during drying. Depending on the scale of the final printed product, different sand fractions can be utilised for optimal results.

All the ingredients were carefully measured and mixed into a uniform mass. Achieving a fine balance among these key ingredients was crucial for a successful print and could only be developed through time-consuming, practice-based research. No artificial hardeners or glues were used during this project, ensuring that all the material is 100% recyclable and reprintable before firing.

From the top: Black Finnish granite sand. Ingredient before mixing in separate buckets and after as a uniform mass in clay mixer.

To determine the optimal firing temperature, test samples were used. These samples illustrated both the color change and the shrinkage in relation to the temperature. Firing at too low a temperature can leave the material too fragile, while firing at too high a temperature can cause deformations and darkening. The ideal temperature yielded a pleasing brick-orange color and provided weatherproof physical properties.

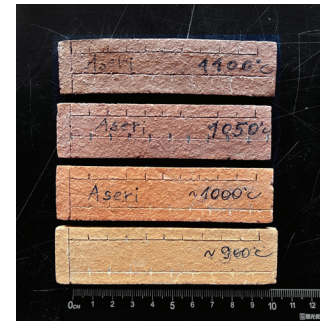

Firing samples.

Some of the objects were tested in harsh weather conditions and have proven their durability. These features bridge the gap between traditional building materials and advanced printing technology, enabling both aesthetic and utilitarian outcomes.

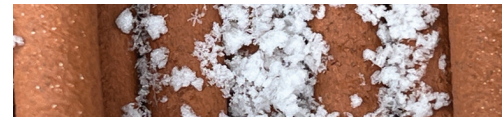

Detail of forzen ceramics.

**Case Study 1: In Conversation by Martin Melioranski**

This object explores the transformation of points into lines and vice versa, as well as the conversion of walls into pillars, through an initial cascading relationship that defines their trajectory in space. The aim is to pose architectural questions about the compression of thought into material lines in space throughout history. The object is scaled at 1:10.

This work builds on close to 24 years of experience with a recursive grammar-based generator known as the Lindenmeyer system (L-system). This system parses a compressed script containing information for movements in space—referred to as axioms and rules—simultaneously into geometric results and "unrolled" interpretations of movements in code. Often described as an Artificial Life system [7] and capable of outputting fractal-like patterns, it relates to Noam Chomsky's work on the classification of languages and formal grammars from the 1950s, where re-writing in productions is applied sequentially. Albeit in L-systems it is done with parallel operations that "…simultaneously replace all letters in a given word," driven by the system's "biological motivations" as described in 1990 by Przemysław Prusinkiewicz and Aristid Lindenmayer in their seminal book "The Algorithmic Beauty of Plants" [8].

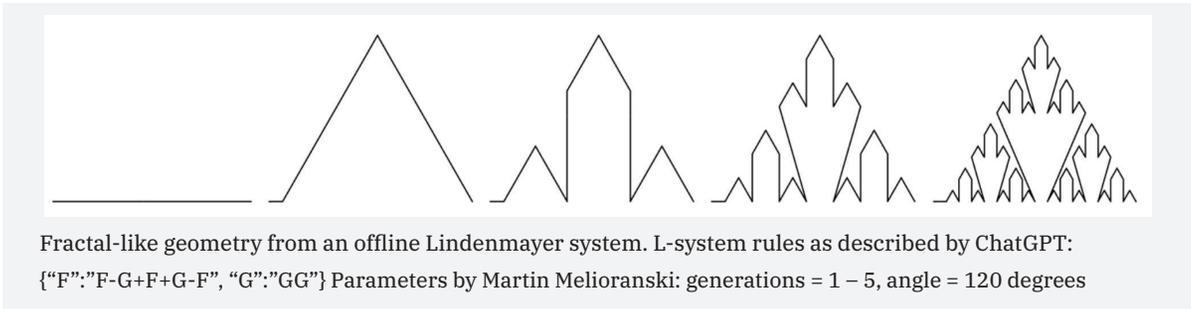

Fractal-like geometry from an offline Lindenmayer system. L-system rules as described by ChatGPT: {"F":"F-G+F+G-F", "G":"GG"} Parameters by Martin Melioranski: generations = 1 – 5, angle = 120 degrees

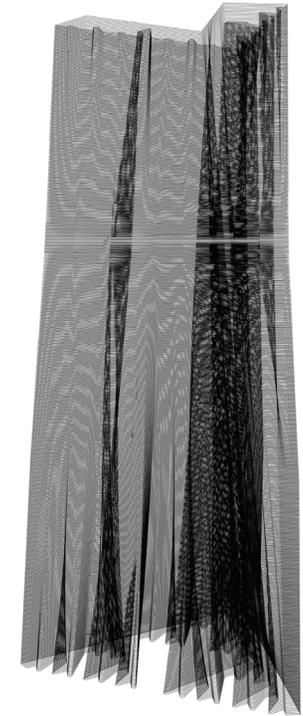

3D view of the robotic toolpath for 3D printing (above) and the continuous NURBS surface that it is derived from (below).

In this case, the initial code was generated through a conversation with ChatGPT 3.5, aiming to create rules for producing a tetrahedron. Although this LLM system failed to move beyond the Euclidean plane, it nonetheless composed an intriguing ruleset. This ruleset was computed into its geometrical output using an offline L-system generator built into Derivative TouchDesigner, and then parsed as XYZ-coordinates into McNeel Grasshopper. By sequencing different generations of the code in the vertical domain to transcend the planar outset, the resulting parametric curves were "manually" combined with direct modeling in the CAD software Rhinoceros 3D to form a continuous NURBS surface.

Top view of the robotic toolpath for 3D printing.

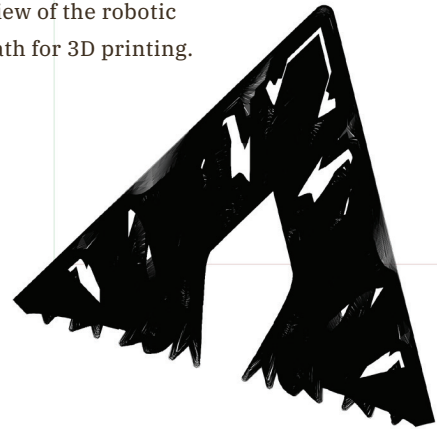
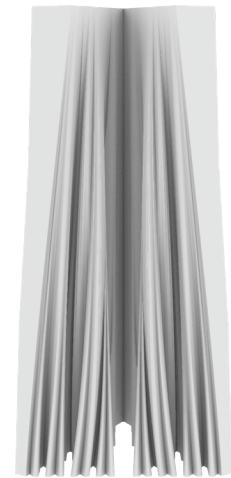

Subsequently, the parametric add-ons of the Rhinoceros platform were used to slice the geometry for 3D printing in clay. The continuous curve was then connected to a UR10 Universal Robot via custom middleware developed at EKA.

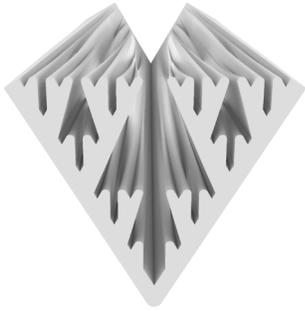

The production of the 3D print introduced additional misalignments to the initial spatial target due to the necessity of remote-controlled operation via AnyDesk. Consequently, when the robotic printing process was unattended between 4 – 6 AM, presumably affected by changes in temperature and humidity, the flow of clay became intermittently disrupted without completely stopping. This resulted in "bird's-nest-like" layers for a brief period, followed by a continuous curvilinear layer, likely due to a more moist mixture of clay. Since the base layer for this part of the sequence was uneven—resembling a "nest"—the subsequent layers adhered to the initial geometry of the toolpath in a more indirect and loose manner. After this smoother phase, the disruptive nest-like structure reappeared, leading to the process being halted before the toolpath reached its final trajectory. The drying process and kiln preserved the raw outcome of these sequences, and the work was exhibited at the museum in its unaltered state.

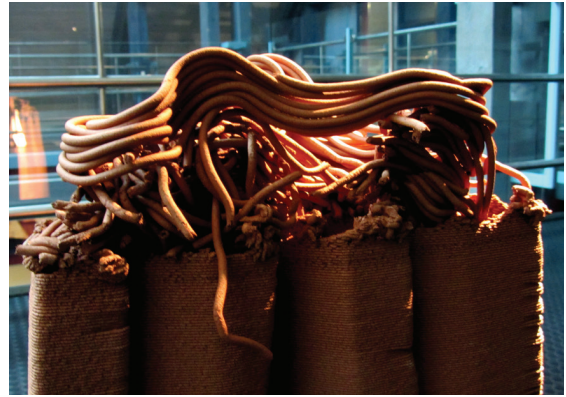

Close-up of the "bird's-nest".

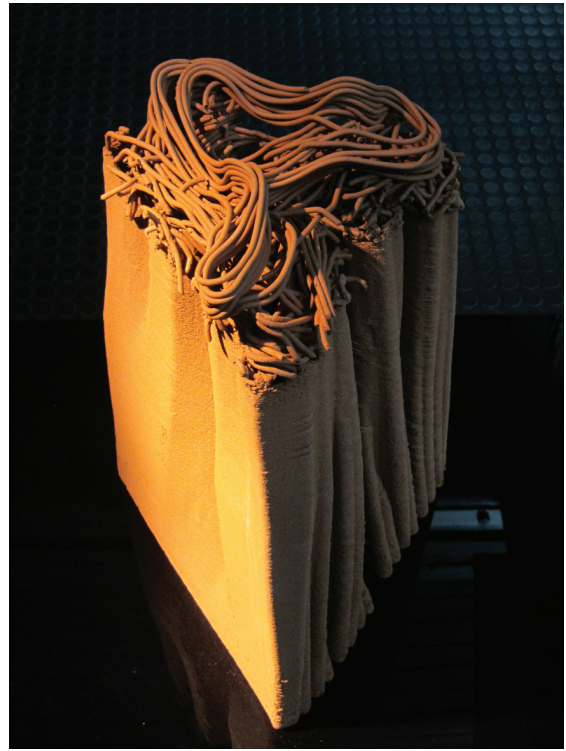

In Conversation as exhibited at Loading Ceramics.
Photo: Juta Kübarsepp

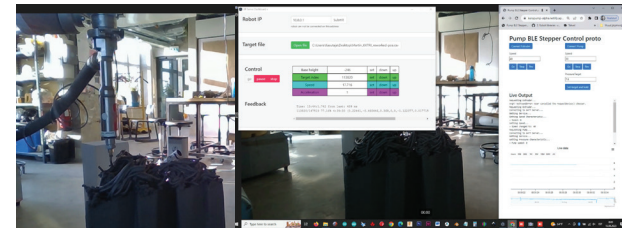

3D clay printing process directed through remote controlled pump settings with AnyDesk.

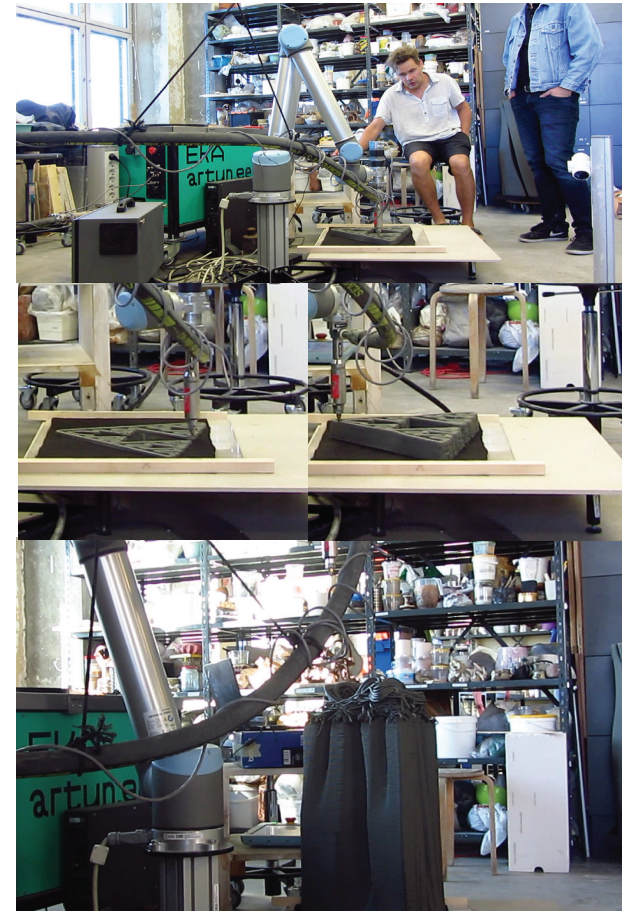

3D clay printing process and final outcome.

**Case Study 2: Braided Path by Lauri Kilusk**

In response to the curatorial statement, the artist explored the intersection of decorative, utilitarian, and structural aspects. Through this somewhat abstract spatial concept, the author encouraged a reconsideration of additive manufacturing methods, opening discussions for further developments rather than providing a final solution.

The design process involved a unique method of multiplying specific points and connecting them with a consistent pattern to create a continuous line. This line then served as a path for a collaborative robot, setting a trajectory of movement without the need for sliced 3D surfaces.

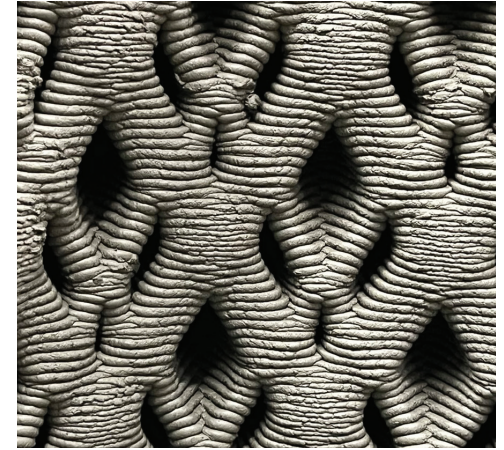

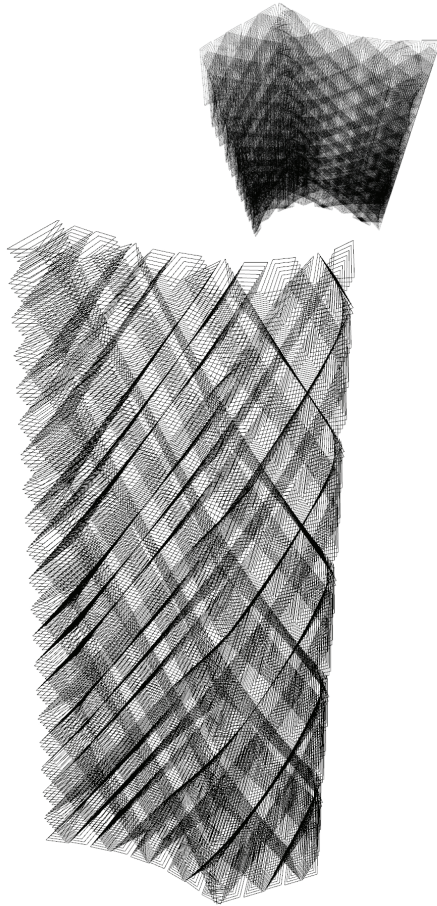

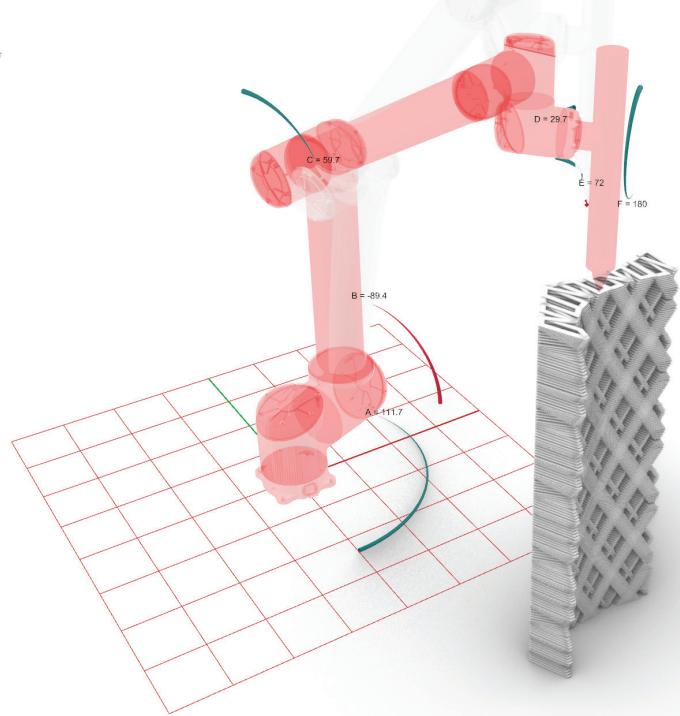

Computer simulation with UR10 Robot.

Screenshots from parametrically controlled 3D-modelling software.

The form as such only emerged in the printing process, when the layers of clay are stacked on top of each other to fill the space between the floating lines. The material revealed the principle of structural integrity and gave the final appearance of intertwined lines.

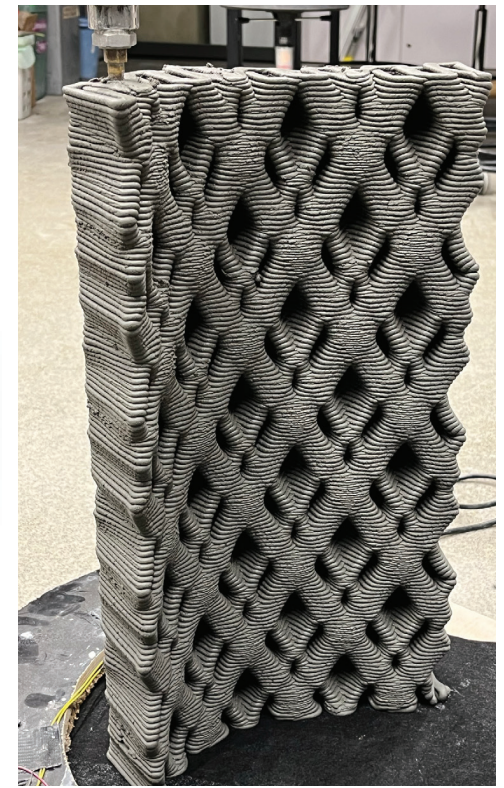

Photos of printed object in its "raw" state.

As the layers of soft clay accumulated, a critical threshold was reached where the weight began to compress the underlying structure. Since clay hardens when exposed to air, the printing speed should not exceed the natural drying process. To ensure stability and prevent deformations, additional supports were added where needed.

The transformation of clay into ceramics through kiln firing endowed the piece with a distinctive orange hue and enhanced water and heat resistance. This experimental process paves the way for further exploration, particularly in leveraging the material's high thermal capacity and porosity for temperature and humidity regulation. Originating from parametric design, this concept offers scalable and adaptable solutions, promising diverse applications across various domains.

The result is a module with an articulated surface.

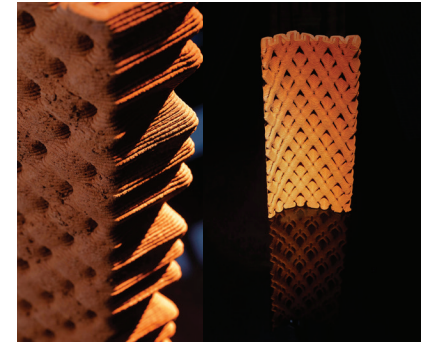

Photo: Juta Kübarsepp

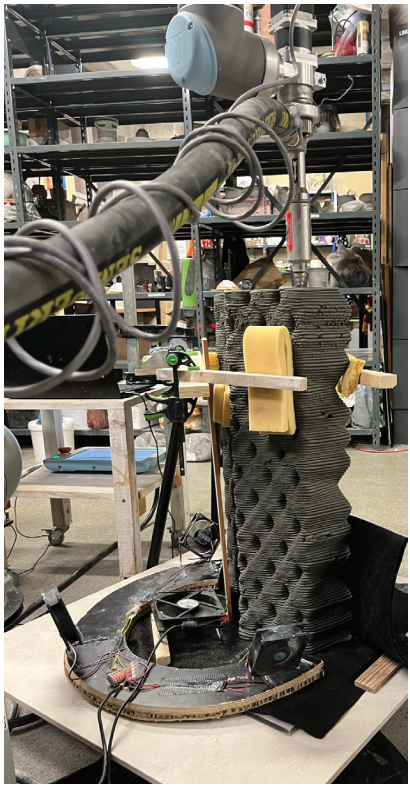

The 3D-printing process with additional support structures.

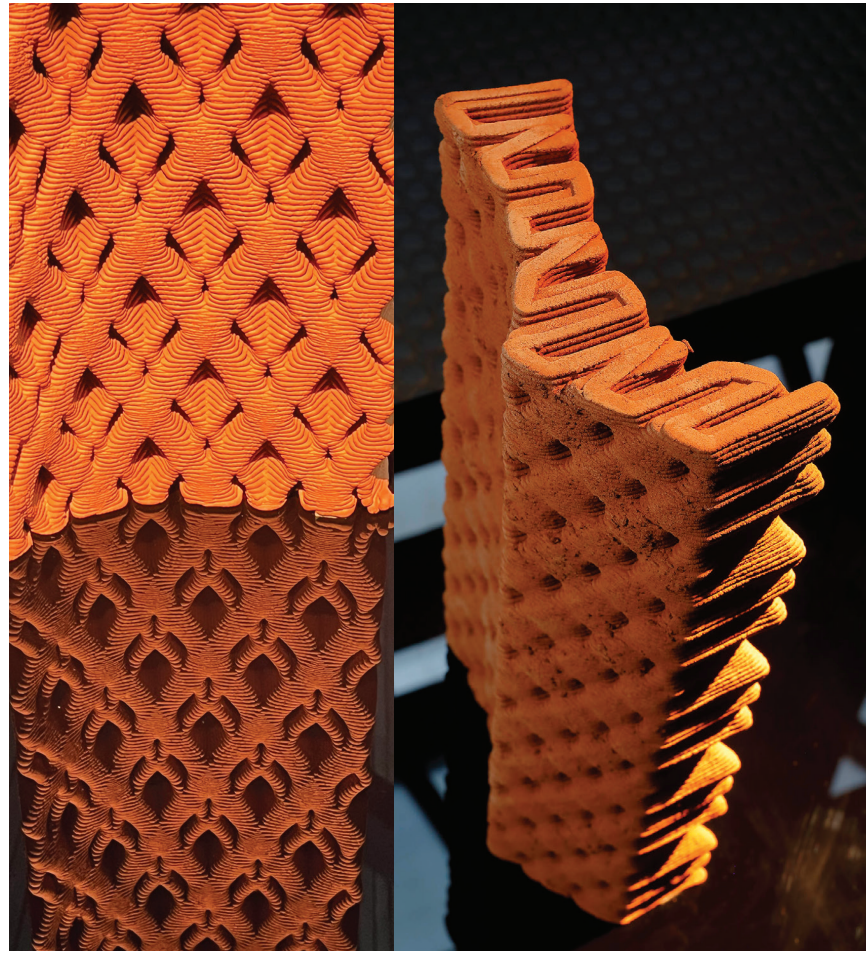

Braided Path, exhibited at Loading Ceramics. Photo: Juta Kübarsepp

Finding the right settings for the print. An attempt to print ovvernight, the result in the morning.

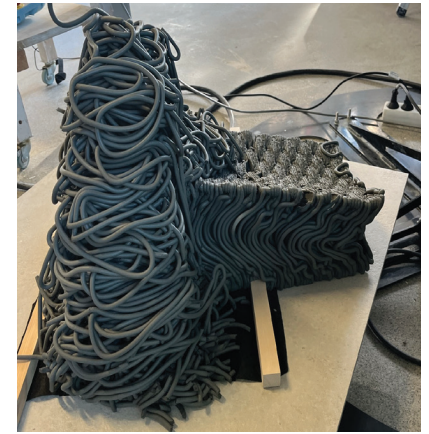

**Case Study 3: Beneath the Cloud by Varvara & Mar (Varvara Guljajeva and Mar Canet Sola)**

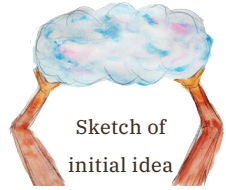

Sketch of initial idea

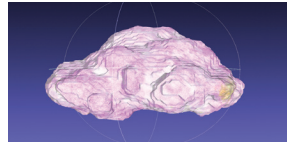

AI-generated 3D model with Dream Fusion (text-to-3D)

In response to the curatorial statement, the artist duo aimed to draw attention to the foundational pillars of today's neural technology. While AI is frequently and excessively hyped as possessing the capability to think, make decisions, and even create art autonomously, it fundamentally relies on human labor. This labor is, at times, carried out under unjust conditions and the human part is never credited. As we comment in one of our previous art projects Keep Smiling (2021) and related paper, decision-making is handed over to machines and labor culture is absurdly based on extraction and monitoring [9].

The artwork, integrating two contrasting components, illustrates this theme. The first element, a set of digitally scanned human hands scaled by two to their actual size, represents the human labor integral to technology. The second, an AI-generated cloud (scale 1:100), symbolizes the expansive nature of AI technology. Previously, the artist duo has utilized AI for 3D mesh stylization as a starting point for AI-infused ceramic sculpture production [10] [11]. This time a text-to-3D DL model was used.

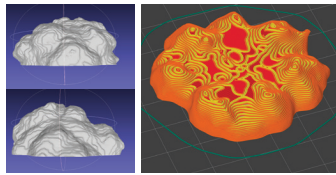

The cloud was divided into two. Slicer view of problematic areas in red.

A noteworthy innovation in this project was the development of an infill feature. Common in FDM 3D printing with plastic, infill is unusual in ceramics due to its weight-increasing effect. Previously, the EKA 3D clay robot had lacked this feature. The residency's outcome was thus twofold according to the methodology explained previously: not only the artworks but also a refined printing technique allowing for more complex, multi-layered forms. A screenshot of the slicer program indicated a problem with the upper layer and a need for infill to fabricate the model successfully. Hence, infill added both stability and weight; in this project, it was strategically placed within 5cm of the sculpture's outer wall, necessitating an internal wall in the digital model to maintain a hollow core and reduce weight. Nonetheless, the final sculptures were hefty, with each cloud side 40cm in diameter and 40cm in height weighing 50kg, totalling 100kg. In contrast, a 75cm tall single-layered hand sculpture was about 10kg.

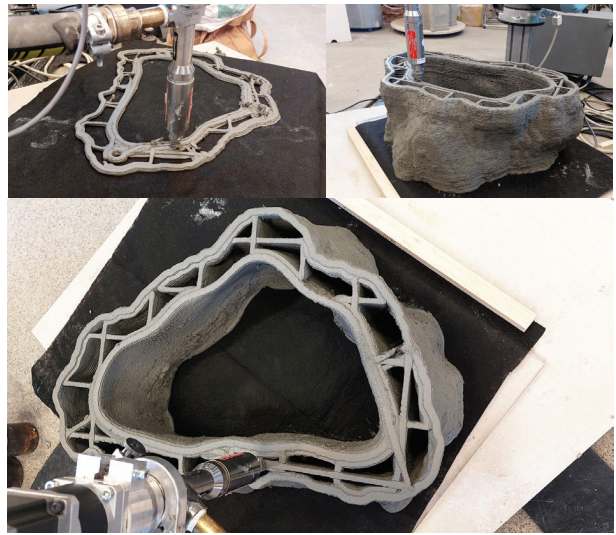

Incorporating sand at the print's base during printing adds weight, preventing collapse as the model's height increases. Internal supports were also utilized to prevent potential deformations.

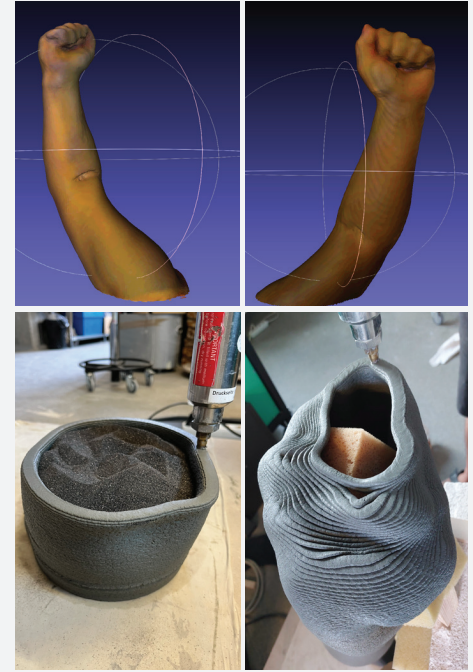

The models with infill differed significantly from those without. The latter required extensive human intervention due to their fragility, as the single-layer clay walls risked collapsing at angles greater than 30 degrees. Artists and architects became craftsmen who had to meticulously identify and reinforce weak spots without marking or causing a collapse of the final clay object, as the accompanying images demonstrate.

Robotic 3D printing process during the first test of the infill.

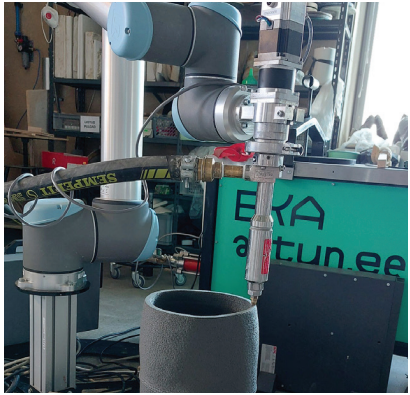 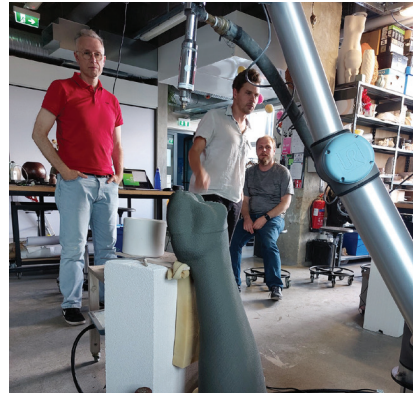 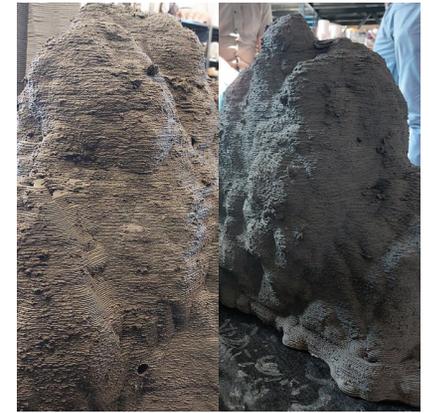

While models with infill require less manual support, they take longer to print. For example, printing a clay hand took the robot 4 hours, whereas a cloud section with infill required 12 hours. The decreased need for manual support led to the decision to permit overnight, unsupervised printing. This process was periodically monitored via webcam through AnyDesk, enabling remote oversight to halt the operation upon completion. The absence of a stop-and-go feature in the clay pump presented a unique challenge. As a result, the pump continued to dispense clay until manually stopped. Fortunately, clay's inherent flexibility while wet proved advantageous; any excess material that had accumulated overnight could be effortlessly removed the next morning.

Created during a five-day artist residency, this work utilized 3D scanning and the Dream Fusion DL model to generate 3D models, which were then clay-printed by a robot. Of the four printed pieces (two hands and two cloud sections), only the cloud was ultimately displayed, featuring neon tubes inside it to juxtapose the natural material with artificial neon light.

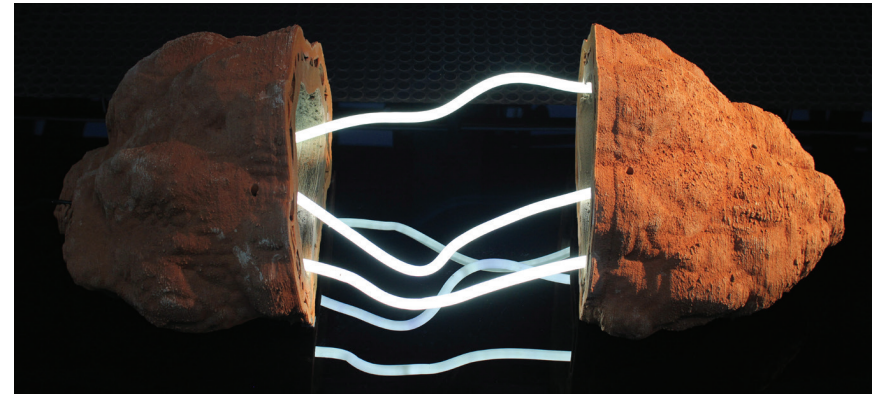

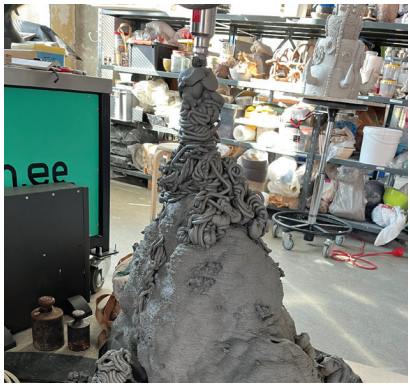

Result of overnight printing showing excess material due to the missing stop-and-go function.

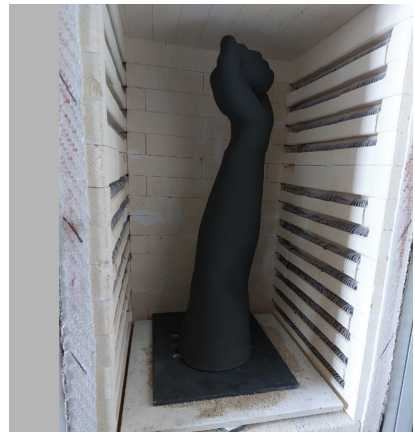

Clay hand in kiln, ready for firing

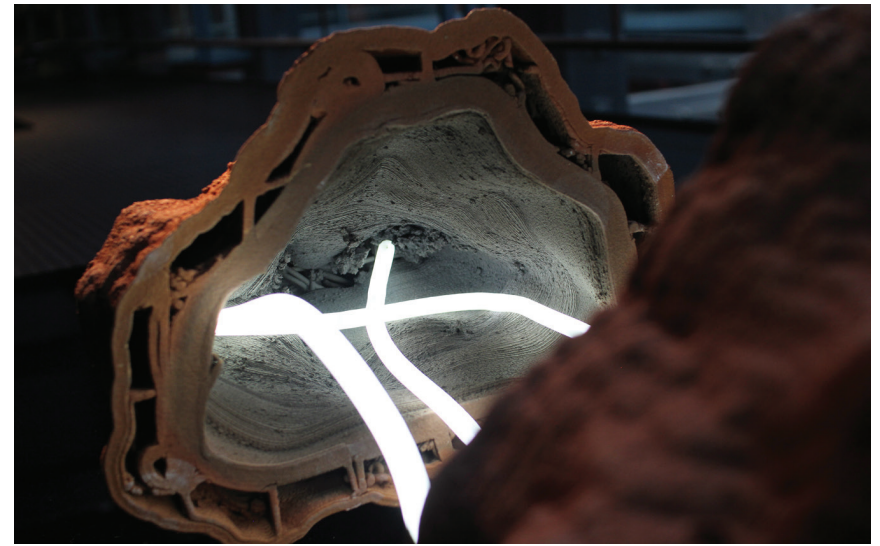

**Case Study 4: Balancing Tower by Kaiko Kivi**

Producing custom ceramics is generally a slow process, involving multiple stages of preparation and post-production to achieve a finished object. The "Balancing Tower" was created as part of development experiments during the residency. Vertically loading wet and plastic clay is a careful balancing act, even with the precision of a computer numerically controlled robot extruder, operating with micrometer and microsecond accuracy. This piece represents a collection of easily buckling, asymmetric forms attempting to support each other and maintain balance as a cohesive structure. The conceptual idea was to provide a glimpse into and through a loaded column of clay. The tower aspires to both resemble and diverge from the typical aesthetics of single-extrusion objects commonly seen in 3D-printed ceramics. Artistic intervention in design is seen as an enabler of value creation in automated production [12]. By leveraging universal production techniques, it becomes possible to shift from the modern, speculative production methods — which aim to meet all conceivable needs and subsequently generate excess — to a more sustainable, on-demand manufacturing paradigm. This approach utilises 3D printing technology to deliver customised designs upon request, significantly reducing the consumption of limited resources [13].

The workflow described here has the aim to coordinate extruder with the robot operation in order to create more complex objects with holes, infills and special structures. The goal of further development would be to develop a seamless workflow for prints that can involve multiple bodies, infill and support structures and more sophisticated coding. The use of common 3D print slicing tools as a basis for robotic clay printing was investigated from ground up as our setup differs from common 3D printing solutions in operation and material deposition.

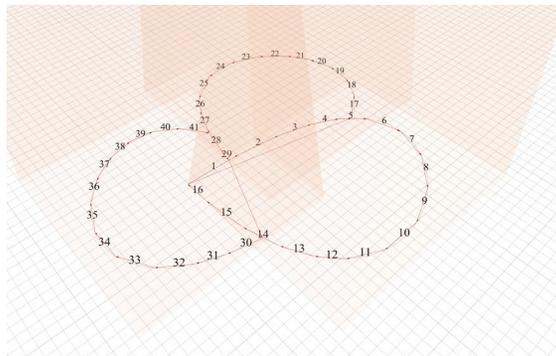

Numbered steps of movement

In general an FDM print job is programmed by splitting the paths into linear steps layer by layer and inserting travel moves between starting and stopping the extruder.

The centre of accumulated mass was then visually studied in digital simulation to balance the tower and ensure better stability during the extrusion. Nevertheless the second attempt of the object collapsed after completion.

The primary objective and a notable development in this project was to synchronize the controlled starting and stopping of the extruder with the print program, and to evaluate the fusion at the seams. To achieve this, the object was designed as a simple repetition of similar layers stacked on top of each other.

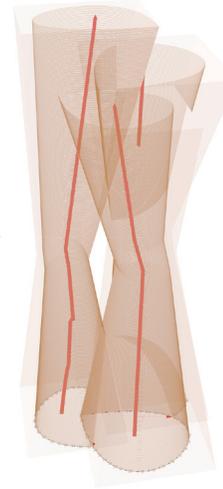

Stacked layers with centers of mass

This design allowed for the observation of quality aspects in the process, enabling adjustments to the program based on our findings. Timing the erection of high structures in plastic clay becomes a negotiation with the structural properties of shape, material and environment. Clay mass hardens while drying which on one hand gives better stability, but it also brings along issues with shrinking, yielding deformations and cracking.

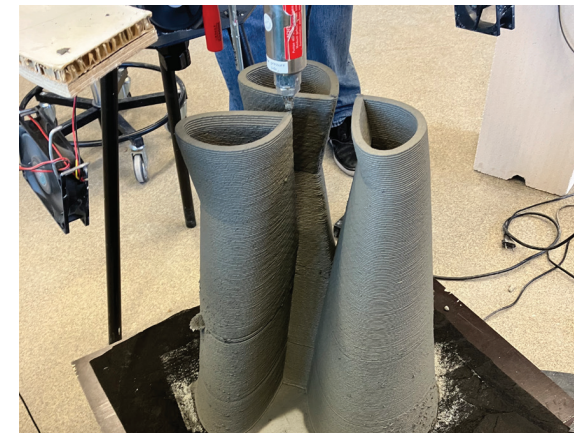

Printing in process with some visible defects.

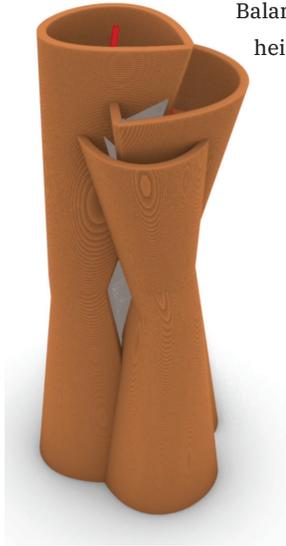

Digital simulation of the Balancing Tower, height 800mm.

Robotic automation provides precision and predictability for the tools' movement and switching the extruder. Every aspect of the automated process needed to be defined upfront. The program for the robot was composed of 55,592 individual movements totalling 682m movement in length, that was executed during 6 hours and 18 minutes.

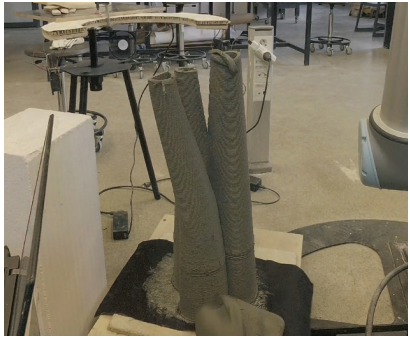

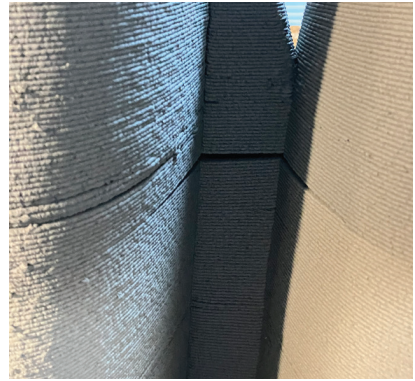

Cracks forming in the first print due to unequal drying.

The robotic system lacks the knowledge of material and object properties making it hard to predict deformations during the operation. Uniform contraction could easily be considered in the program, but drying and deformations of the raw object depend on the shape, environmental conditions and time that make it quite infeasible in custom production settings. The object is also often deforming in unpredictable ways, which can prove difficult to account for in programming of custom objects. Unlike typical 3D printing with fairly predictable materials such as plastic polymers, clay often requires sensitive control, making it a live and creative process.

Collapse of the second attempt.

The material itself allows ways of manual intervention and the slow extrusion process allows for tactile human assistance augmenting the lacking sensibility of the robotic system. The ceramics printing process developed at EKA has continuously incorporated real-time control and manual intervention throughout both the digital design

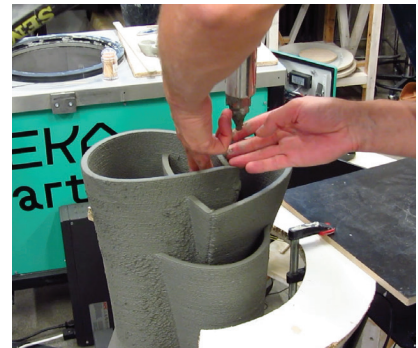

Amending the print after a program bug crashed the extruder into it.

and robotic production stages, often ahead of automation and during post-production. Human control allows for greater versatility and artistic involvement in the production process, enabling creative expression and the ability to amend mistakes in later phases. Experiments have shown that the extruder can be finely controlled in unison with the robot, unlocking significant creative potential.

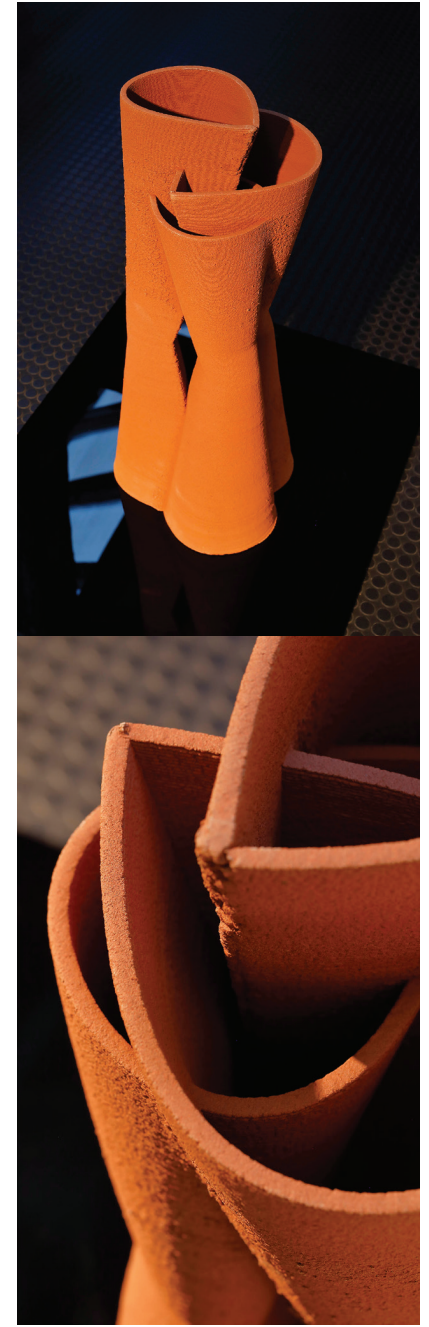

Photos by Juta Kübarsepp

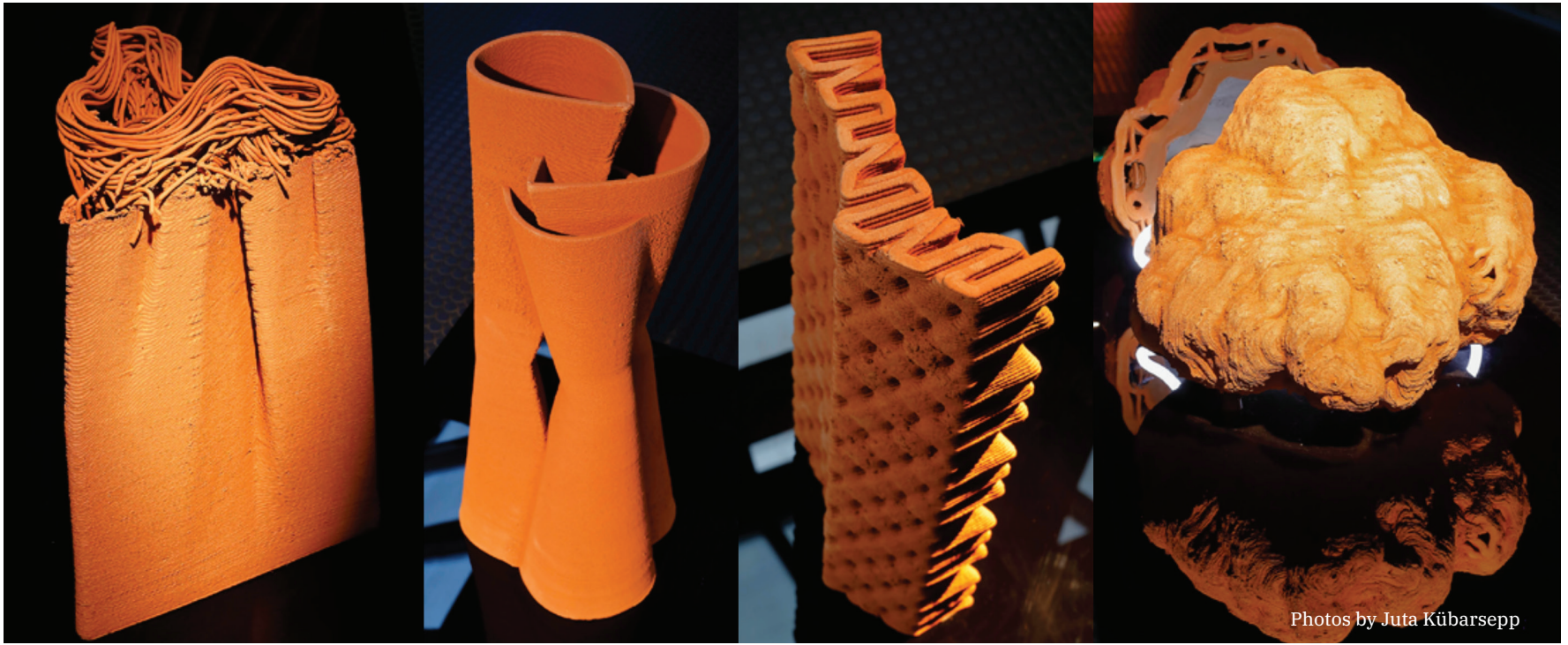

Photos by Juta Kübarsepp

**Discussions and Conclusions**

The integration of parametric digital tools with materials like clay and various biomaterials, coupled with the historical nuances of creative fields and the so-called tacit knowledge [14], has necessitated innovation on both personal and global scales. Progress in creative fields requires advancements in individual creations, technological developments, and more broadly on a social level [15]. This project's multidisciplinary and hybrid nature underscores its significance; without the cooperation across different disciplines, the results achieved thus far would not have been possible. However, it is evident that the accomplishments to date are merely a foundation for future advancements. To achieve even better results, precise technological development must be accompanied by the promotion of co-design, efficient information management solutions, and clear communication of results through various mediums, including text, diagrams, photos, and animation-videos.

This project has yielded significant outcomes, marking a milestone in the field of ceramic 3D printing. Notably, it has produced the largest 3D printed ceramic objects in Estonia, with notable international relevance when compared to some of the latest developments [16]. This achievement not only showcases the technical capabilities of the project's bespoke machinery and software but also highlights the potential of 3D printing technology to reach new heights in artistic and architectural fabrication.

A critical aspect of this project was its innovative approach to material usage and fabrication processes, such as infill strategies and the implementation of stop-and-go techniques with custom-made fabrication equipment. These methodologies have allowed for greater efficiency and flexibility in the production of ceramic objects, addressing common challenges like drying time and material shrinkage.

The project also revealed unique challenges associated with unsupervised printing, particularly in managing errors and ensuring quality control. The ability to monitor and adjust the printing process remotely has opened up new possibilities for ceramics fabrication but also underscored the need for robust error detection and mitigation strategies.

One of the most distinctive contributions of the Loading Ceramics initiative is its development of a unique hybrid research methodology that seamlessly integrates RtD and artistic research. This methodology has facilitated a rich, multidimensional exploration of 3D printing in clay, allowing for a holistic examination of both the process and the final artefacts. By combining these approaches, the project has fostered an environment where technical and material innovation coexist with artistic expression, leading to meaningful results that push the boundaries of traditional practices. This hybrid research methodology not only amplifies the depth and breadth of inquiry but also encourages a more collaborative and interdisciplinary approach to research and development in the field. It underscores the value of combining technical and creative perspectives to address complex challenges. Moreover, this research has helped to visualize the material and structural possibilities of robotics in ceramics. By demonstrating how robotic technology can be utilized in the fabrication of complex ceramic structures, the project provides insights into new ways of thinking about material properties and design potential.

In conclusion, through this research project, participants were empowered to experiment with and refine the robotic-ceramics interface, leading to the production of experimental ceramic columns and pillars. These artefacts not only challenge traditional notions of support structures in architecture but also serve as a testament to the project's success in fostering multidisciplinary collaboration and technological advancement.

Furthermore, the research contributes towards the integration of robotics into ceramics, showcasing an innovative blend of art, architecture, and practice-based research. This interdisciplinary endeavour has not only fostered a deep exploration of the material and technological facets of ceramic 3D printing but also ignited a collaborative spirit among participants, leading to significant advancements in both the creative and technical domains. The establishment of a specialised in-house smart extruder and the development of custom software for controlling the clay printing machinery have been pivotal in overcoming the inherent limitations of existing 3D clay printers. These technical innovations have enabled the project to push the boundaries of what's possible in ceramic fabrication, offering a glimpse into the future of architectural and artistic expression.

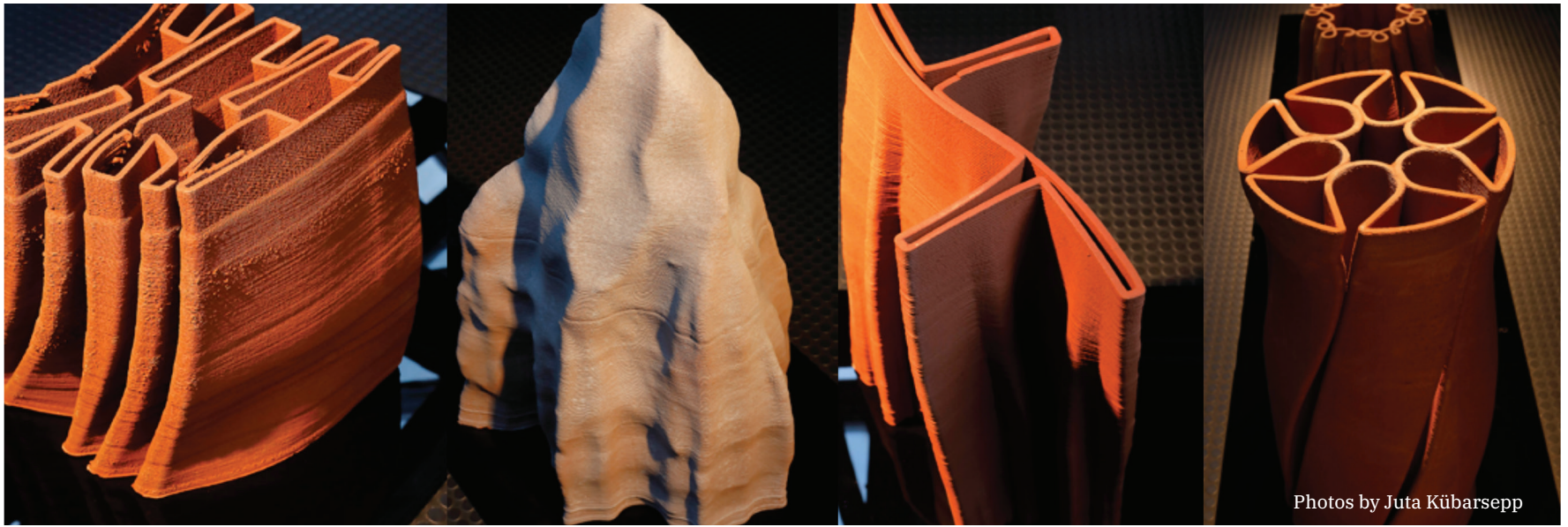

Photos by Juta Kübarsepp